\documentclass[11pt,a4paper,dvipsnames]{article}
\usepackage[hyperref]{acl2022}
\usepackage{times}
\usepackage{latexsym}

\usepackage{microtype}

\aclfinalcopy 

\usepackage{multirow}
\usepackage{enumerate}
\usepackage{booktabs}
\usepackage{adjustbox}
\usepackage{amsmath}
\usepackage{soul}
\usepackage{enumitem}

\title{MIMICause: Representation and automatic extraction of causal relation types from clinical notes}

\author{\textbf{Vivek Khetan}\thanks{~ Corresponding Author}\phantom{\footnotesize 1}\thanks{~ Equal contribution}$^{~\spadesuit}$
\quad \textbf{Md Imbesat Hassan Rizvi}\footnotemark[2]$^{~~\clubsuit}$
\quad \textbf{Jessica Huber}\thanks{~ Contributed during an internship at Accenture labs, SF}\phantom{\footnotesize 1}$^{~\heartsuit}$ \\ 
\smallskip 
\quad \textbf{Paige Bartusiak}\footnotemark[3]\phantom{\footnotesize 1}$^{~\diamondsuit}$ 
\quad \textbf{Bogdan Sacaleanu}$^{\spadesuit}$ 
\quad \textbf{Andrew Fano }$^{\spadesuit}$ \\

  $^\spadesuit$Accenture Labs, SF,   $~~^\clubsuit$Indian Institute of Science\\ 
  $^\heartsuit$Arizona State University, $~~^\diamondsuit$Duke University \\
  
  \texttt{vivek.a.khetan@accenture.com}, \texttt{mdrizvi@iisc.ac.in}\\ \texttt{jhuber12@asu.edu}, \texttt{paige.bartusiak@cs.duke.edu}\\ \texttt{\{bogdan.e.sacaleanu, andrew.e.fano\}@accenture.com} \\
  \\}

\date{March 2022}

\begin{document}
\maketitle

\begin{abstract}
Understanding causal narratives communicated in clinical notes can help make strides towards personalized healthcare. Extracted causal information from clinical notes can be combined with structured EHR data such as patients' demographics, diagnoses, and medications. This will enhance healthcare providers' ability to identify aspects of a patient's story communicated in the clinical notes and help make more informed decisions.  

In this work, we propose annotation guidelines, develop an annotated corpus\footnote{MIMICause dataset will be available under the ``Community Annotations Downloads'' at \href{https://portal.dbmi.hms.harvard.edu/projects/n2c2-nlp/}{https://portal.dbmi.hms.harvard.edu/projects/n2c2-nlp/}} and provide baseline scores to identify types and direction of causal relations between a pair of biomedical concepts in clinical notes; communicated implicitly or explicitly, identified either in a single sentence or across multiple sentences. 

We annotate a total of 2714 de-identified examples sampled from the 2018 n2c2 shared task dataset and train four different language model based architectures. Annotation based on our guidelines achieved a high inter-annotator agreement i.e. Fleiss' kappa ($\kappa$) score of 0.72, and our model for identification of causal relations achieved a macro F1 score of 0.56 on the test data. The high inter-annotator agreement for clinical text shows the quality of our annotation guidelines while the provided baseline F1 score sets the direction for future research towards understanding narratives in clinical texts. 
\end{abstract}
\section{Introduction}

Electronic Health Records (EHRs) have significant amounts of unstructured clinical notes containing a rich description of patients' states as observed by healthcare professionals over time. Our ability to effectively parse and understand clinical narratives depends upon the quality of extracted biomedical concepts and semantic relations. 
The contemporary advancements in natural language processing (NLP) have led to an increased interest in tasks such as extraction of biomedical concepts, patients' data de-identification, medical question answering and relation extraction. While these tasks have improved our ability for clinical narrative understanding, identification of semantic causal relations between biomedical entities will further enhance it. 


Identification of novel and interesting causal observations from clinical notes can be instrumental to a better understanding of patients' health. It can also help us identify potential causes of diseases and determine their prevention and treatment.  
Despite the usefulness of identification and extraction of causal relation types, our capability to do so is limited and remains a challenge for specialized domains like healthcare. 

\begin{figure*}[!th]
\centering
\begin{minipage}{1.0\textwidth}
  \centering
  \includegraphics[width=\linewidth,keepaspectratio]{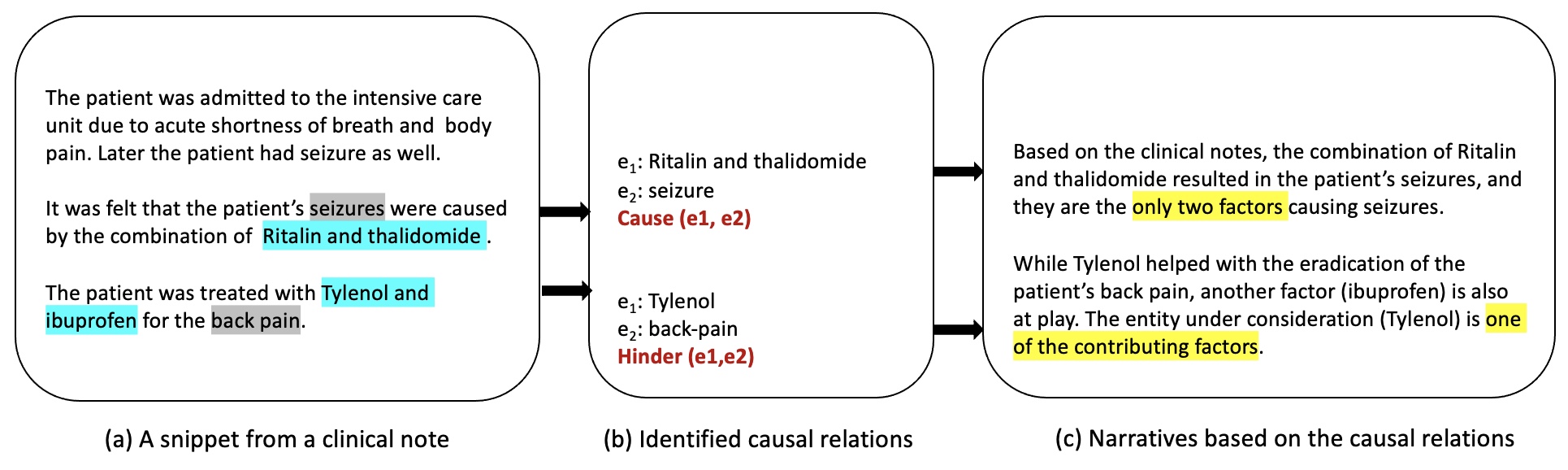}
  \caption{(a) A snippet from a clinical note with highlighted biomedical entities identified in the n2c2 dataset. (b) Causal relations identified between the specified biomedical entities (e1 and e2). In the first case, two entities are specified together as e1 for causal relation identification, while the second case specifies only one entity as e1. (c) Narratives based on the causal relations identified between the specified biomedical entities}
  \label{fig:MIMICause-overview}
\end{minipage}%

\end{figure*}

The NLP community has been actively working on causality understanding from text and has proposed various methodologies to represent  \cite{ForceDynamics, dynamic-causal-model,necessary-and-sufficient,Hassanzadeh2019AnsweringBC}, as well as extract \cite{Mirza2014AnAO, OGorman2016RicherED, mirza-tonelli-2016-catena,Gao2019ModelingDC, 10.1007/978-3-030-80119-9_64}, causal associations between the events expressed in natural language text. 
In the healthcare domain, most of the related work can be grouped around the problem of adverse drug effect identification from biomedical scientific articles \cite{Gurulingappa2012DevelopmentOA} or clinical notes \cite{Johnson2016MIMICIIIAF, liu2019towards,Henry20202018NS, rawat2020inferring}, and identification of cause, effect and their triggers \cite{Mihaila2012BioCauseAA}. There is no work that has yet tried to represent different types of causal associations along with direction (between biomedical concepts) communicated in clinical notes.

In this work, we fill the gap by defining types of semantic causal relations between biomedical entities, building detailed annotation guidelines and annotating a large dataset. 

Figure \ref{fig:MIMICause-overview} shows a snippet of clinical note extracted from the n2c2 dataset \cite{Henry20202018NS}, different sets of annotated biomedical entities along with the causal relationship between them, and the corresponding narrative based on the proposed guidelines outlined in Section \ref{sec:guidelines}. 


Even with the inherent complexities of clinical text data (e.g., domain knowledge, short hand by doctors, etc.), following our proposed guidelines, we achieved a high inter annotator agreement of Fleiss' kappa ($\kappa$) score of 0.72.

\section{Related Works}
In linguistics, the focus on representing causality has been on understanding interactions between events. Talmy (\citeyear{ForceDynamics}) proposed force-dynamics to decompose the causal interaction between events as ``letting'', ``helping'', ``hindering'' etc. Wolff (\citeyear{dynamic-causal-model}) built upon force-dynamics by incorporating the theory of causal verbs and proposed the Dynamic-model of causation. Wolff categorised causation in three categories, \textit{``Cause''}, \textit{``Enable''} and \textit{``Prevent''}, and provided a set of causal verbs to express these categories.  

Dunietz et al. (\citeyear{Dunietz2015AnnotatingCL, Dunietz2017TheBC}) proposed BECauSE Corpus to represent linguistic expressions of causation stated explicitly. BECauSE 1.0 \cite{Dunietz2015AnnotatingCL} consists of a cause span, an effect span, and a causal connective span. Their work treats the causal connectives e.g. \textit{because of, so} etc. as the ``centerpiece'' of causal language, impacting the selection of instances to be annotated. In addition to the types of causation (Consequence, Motivation, and Purpose) and degrees of causation (Facilitate and Inhibit) introduced in BECauSE 1.0, the subsequent work BECauSE 2.0 \cite{Dunietz2017TheBC} extended the annotation scheme to include overlapping relations other than causal. In contrast, our work focuses on both explicit (indicated by connectives) and implicit (lack of connectives) identification of types of causal associations between biomedical concepts as communicated in clinical notes.

More recently, Mostafazadeh et al. (\citeyear{Mostafazadeh2016CaTeRSCA}) built upon the work of Wolff and proposed annotation framework \textit{CaTeRS} to represent causal relations between events for commonsense perspective. CaTeRS categorises semantic relations between events to capture causal and temporal relationships for narrative understanding on crowd-sourced ROCStories dataset \cite{Mostafazadeh2016ROCStories} but has only 488 causal links. In comparison, our MIMICause dataset is built on actual clinical narratives, i.e., \textit{MIMIC-III Clinical text data} \cite{Johnson2016MIMICIIIAF} and has 1923 causal observations.

Another interesting decomposition of causation is proposed by Swartz (\citeyear{necessary-and-sufficient}) as a necessary and sufficient condition, but such detailed information is seldom communicated in clinical notes. There have been several other recent attempts of modeling and extracting causality from unstructured text. Bethard et al. (\citeyear{Bethard2008LearningSL}) created a causality dataset using the Wall Street Journal corpus and captured the directionality of causal interaction with simple temporal relations (e.g., Before, After, No-Rel) but did not focus on the types of causality between the events. The work of Gorman et al. on Richer Event Description (RED) \cite{Ikuta2014ChallengesOA} describes causality types as cause and precondition and uses negative polarity to capture the context of hinder and prevent. This is in line with the annotation guidelines proposed in our current work, but we also defined explicit \textbf{Hinder} and \textbf{Prevent} causality types along with directionality. 

Mirza et al. (\citeyear{Mirza2014AnAO}) proposed the use of explicit linguistic markers, i.e., CLINKs (due to, because of, etc.) to extended TimeML TLINKs \cite{Pustejovsky2003TimeMLRS} based temporal annotations to capture causality between identified events. The resulting dataset had temporal as well as casual relations but still lacks the causality types between events. Hassanzadeh et al. (\citeyear{Hassanzadeh2019AnsweringBC}) proposed the use of binary questions to extract causal knowledge from unstructured text data but did not focus on types and directionality of causal relations. More recently, Khetan et al. (\citeyear{10.1007/978-3-030-80119-9_64}) used language models combining event descriptions with events' contexts to predict causal relationships. Their network architecture wasn't trained to predict the type or directionality of causal relations. Furthermore, they removed the directionality provided in SemEval-2007 \cite{Girju2007SemEval2007T0}, and SemEval-2010 \cite{Hendrickx2009SemEval2010T8} datasets to evaluate their model on a larger causal relation dataset. Our causality extraction network is built upon their methodology, i.e., \textit{Causal-BERT} but also focuses on directionality as well as types of causality communicated in clinical notes.

Although causality lies at the heart of biomedical knowledge, there are only a handful of works (mostly Adverse Drug Effect (e.g. Gurulingappa et al. \citeyear{Gurulingappa2012DevelopmentOA})) extracting causality from biomedical or clinical text data. 
Uzuner et al. (\citeyear{Uzuner20112010IC}) proposed tasks to extract concepts, assertions, and relations in clinical text. In their dataset, drugs and procedures are combined as a single concept, i.e., treatment and the defined relations are also dependent upon the concept types under consideration. Whereas, the relations defined in our work are based on the overall context in any given example and make no assumption about the type of concepts/entities under consideration.    

Another interesting work is BioCause by Mihaila et al. (\citeyear{Mihaila2012BioCauseAA}), which annotates existing bio-event corpora from biomedical scientific articles to capture biomedical causality. Instead of identifying the types (and direction) of causal relations in the already provided events of interest, they are annotating two types of text spans, i.e., arguments and triggers. Arguments are text spans that can be represented as events with type Cause, Effect, and Evidence while Trigger spans (can be empty) are connectives between the casual events.

Our work proposes comprehensive guidelines to represent the types and direction of causal associations between biomedical entities, expressed explicitly or implicitly in the same or multiple sentences in clinical notes, and is not covered by any related work. 
\section{MIMICause Dataset creation}
We used publicly available 2018 n2c2 shared task \cite{Henry20202018NS} dataset on adverse drug events and medication extraction to build the MIMICause dataset. The n2c2 dataset was used because it is built upon the de-identified discharge summaries from the MIMIC-III clinical care database \cite{Johnson2016MIMICIIIAF} and has nine different annotations of biomedical entities e.g. Drug, Dose, ADE, Reason, Route etc. The types of biomedical concepts/entities with a few examples as defined in the n2c2 dataset are shown in Table \ref{tab:concepts_entities}. 

However, the provided relationships in the n2c2 dataset are simply defined by the identified concepts linked with related medications and hold no semantic meaning. To create the MIMICause dataset, we extracted\footnote{We used \url{https://spacy.io/} library with \textit{``en\_core\_web\_sm''} language model.} examples from each entity-pair available in the n2c2 dataset. Our final dataset has 1107 ``ADE-Drug'' , 1007 ``Reason-Drug''  and 100 from each of ``Strength-Drug'', ``Form-Drug'', ``Dosage-Drug'', ``Frequency-Drug'', ``Route-Drug'' and ``Duration-Drug'' entity-pair examples.

\begin{table}
    \centering
    \begin{adjustbox}{max width=0.48\textwidth}
    \begin{tabular}{c c}
    \toprule
        Concepts/Entities & Examples \\
    \midrule
        Drug & \begin{minipage}{0.48\textwidth} morphine, ibuprofen, antibiotics (or ``abx'' as its abbreviation), chemotherapy etc. \end{minipage} \\
    \midrule
        ADE and Reason$^*$ & \begin{minipage}{0.48\textwidth} nausea, seizures, Vitamin K deficiency, cardiac event during induction etc. \end{minipage} \\
    \midrule
        Strength & \begin{minipage}{0.48\textwidth} 10 mg, 60 mg/0.6 mL, 250/50 (e.g. as in Advair 250/50), 20 mEq, 0.083\% etc. \end{minipage} \\
    \midrule
        Form & \begin{minipage}{0.48\textwidth} Capsule, syringe, tablet, nebulizer, appl (abbreviation for apply topical) etc. \end{minipage}\\
    \midrule
        Dosage & \begin{minipage}{0.48\textwidth} Two (2) units, one (1) mL, max dose, bolus, stress dose, taper etc. \end{minipage}\\
    \midrule
        Frequency & \begin{minipage}{0.48\textwidth} Daily, twice a day, Q4H (every 4 Hrs), prn (pro re nata i.e as needed) etc. \end{minipage}\\
    \midrule
        Route & \begin{minipage}{0.48\textwidth} Transfusion, oral, gtt (guttae i.e. by drops), inhalation IV (i.e. Intravenous) etc. \end{minipage} \\
    \midrule
        Duration & \begin{minipage}{0.48\textwidth} For 10 days, chronic, 2 cycles, over 6 hours, for a week etc. \end{minipage}\\
    \bottomrule
        \multicolumn{2}{l}{\begin{minipage}{0.7\textwidth} $^*$\footnotesize{The distinction between ADE and Reason concepts is based on whether the drug was given to address the disease (Reason) or led to the disease (ADE).} \end{minipage}}
    \end{tabular}
    \end{adjustbox}
    \caption{Examples of Bio-medical concepts/entities in the 2018 n2c2 shared task dataset.}
    \label{tab:concepts_entities}
\end{table}


\subsection{Annotation guidelines}
\label{sec:guidelines}

Our annotation guidelines are defined to represent nine semantic causal relationships between biomedical concepts/entities in clinical notes. Our guidelines have four types of causal associations, each with two directions, and a non-causal ``Other'' class. Based on our guidelines, \textit{causal} relationship/association exists when one or more entities affect another set of entities. The driving concept  can be a \textit{\textbf{single}} entity such as a drug / procedure / therapy or a \textit{\textbf{composite entity}} such as several drugs / procedures / therapies considered together. 

\subsubsection{Direction of causal association}
The direction of causal association between entities is captured by the order of entity tags (($e_1,e_2$) or ($e_2,e_1$)) in the defined causal relationships. Either entity can be referred to as $e_1$ or $e_2$. The \textit{entity that initiates or drives the causal interaction is placed first} in parenthesis followed by the resulting entity or effect.

\begin{enumerate}
        \item \begin{minipage}[t]{0.38\textwidth}Odynophagia: Was presumed due to \textcolor{blue}{\textbf{\textless e2\textgreater mucositis\textless /e2\textgreater}} from recent \textcolor{red}{\textbf{\textless e1\textgreater chemotherapy\textless /e1\textgreater}}.\end{minipage}
        
        \item \begin{minipage}[t]{0.38\textwidth}Odynophagia: Was presumed due to \textcolor{red}{\textbf{\textless e1\textgreater mucositis\textless /e1\textgreater}} from recent \textcolor{blue}{\textbf{\textless e2\textgreater chemotherapy\textless /e2\textgreater}}.\end{minipage}
    \end{enumerate}
Example (1) and (2) are different because the entity references are reversed. Regardless of the entity tags, in the context of the example, ``chemotherapy'' is the \textit{driving entity} that led to the \textit{emergence} of ``mucositis''. Therefore, example (1) is annotated with causal direction ($e_1,e_2$) while example (2) is annotated with ($e_2,e_1$).

\subsubsection{Explicitness / Implicitness of the causal indication}
Our guidelines also capture causality expressed both explicitly and implicitly. In example (1), the causality is expressed explicitly using lexical causal connective ``due to''. Whereas in example (3), the causal association between ``erythema'' and ``Dilantin'' can only be understood based on the overall context of all the sentences.

\begin{enumerate}[resume]
    \item \begin{minipage}[t]{0.38\textwidth} patient's wife noticed \textcolor{blue}{\textbf{\textless e2\textgreater erythema on patient's face\textless /e2\textgreater}}.  On [**3-27**]the visiting nurse [**First Name (Titles) 8706**][**Last Name (Titles)11282**]of a rash on his arms as well.  The patient was noted to be febrile and was admitted to the [**Company 191**] Firm.  In the EW, patient's \textcolor{red}{\textbf{\textless e1\textgreater Dilantin\textless /e1\textgreater}} was discontinued and he was given Tegretol instead. \end{minipage}
\end{enumerate}

\subsubsection{(Un)-certainty of causal association}
 
Establishing real-world causality or the task of causal inference is not in the scope of our current work.  Our proposed guidelines represent a potential causal association between biomedical entities either expressed as speculation or with certainty in a similar manner. 
 \begin{enumerate}[resume]
    \item \begin{minipage}[t]{0.38\textwidth} \textcolor{red}{\textbf{\textless e1\textgreater Normocytic Anemia\textless /e1\textgreater}} - Was 32.8 at OSH; after receiving fluids HCT has fallen further to 30. Baseline is 35 - 40. Not clinically bleeding. Perhaps due to \textcolor{blue}{\textbf{\textless e2\textgreater chemotherapy\textless /e2\textgreater}}. \end{minipage}
\end{enumerate}
In example (4), causality between biomedical entities is speculated through ``Perhaps''. While representing speculative causal associations can further enrich narrative understanding; it is not covered in our current work. 

\subsubsection{Types of causal associations}
This section provides detailed guidelines for various types of causal relations (each with two directions) and one non-causal relation (``Other'') along with accompanying examples.
\begin{itemize}
    \item \textbf{Cause($e_1,e_2$)} or \textbf{Cause($e_2,e_1$)} -- Causal relations between biomedical entities are of these classes if the emergence, application or increase of a \textbf{single or composite entity} \textit{exclusively leads} to the \textbf{emergence} or \textbf{increase} of one or a set of entities. 
\begin{enumerate}[resume]
        \item \begin{minipage}[t]{0.38\textwidth} It was felt that the patient's \textcolor{blue}{\textbf{\textless e2\textgreater seizures\textless /e2\textgreater}} were caused by the combination of \textcolor{red}{\textbf{\textless e1\textgreater Ritalin and thalidomide\textless /e1\textgreater}}. \end{minipage}
    \end{enumerate}
In example (5), ``seizures'' occurred due to \textit{two drugs} viz. ``Ritalin'' and ``thalidomide''. The entity span covers both of them, and they are considered together as a \textit{composite entity} leading to ``seizures''. Hence, example (5) is annotated as Cause($e_1,e_2$). The annotation would have been different had these entities been considered individually.

Thus, the ``Cause'' category is assigned only if the driving entity is responsible in its entirety for the effect. If the specified entity is responsible for the effect in part, then a different causal relation is defined to express this contrast. 

\item \textbf{Enable($e_1,e_2$)} or \textbf{Enable($e_2,e_1$)} -- Causal relations between biomedical entities are of these classes if the emergence, application or increase of a \textbf{single or composite entity} \textit{leads} to the \textbf{emergence} or \textbf{increase} of one or a set of entities in a setting \textit{\textbf{where}} a number of factors are at play and the \textit{single or composite entity} under consideration is one of the contributing factors. 
\begin{enumerate}[resume]
        \item
        \begin{minipage}[t]{0.38\textwidth} It was felt that the patient's \textcolor{blue}{\textbf{\textless e2\textgreater seizures\textless /e2\textgreater}} were caused by the combination of \textcolor{red}{\textbf{\textless e1\textgreater Ritalin\textless /e1\textgreater}} and thalidomide. \end{minipage}
        
    \end{enumerate}
Example (6) is the same as example (5) except for the entities in considerations. Both the drugs viz. ``Ritalin'' and ``thalidomide'' are contributing to the ``seizures''.

Since the example is considering only ``Ritalin'', \textit{ which is a contributing factor in part}, it is annotated as Enable($e_1, e_2$). 

With the ``Enable'' relation type, it can easily be noted that discontinuing only ``Ritalin'' or ``thalidomide'' will not lead to the stopping of ``seizures''. Labelling these samples as ``Cause'' would have suppressed this detail, and the actions taken based on this would not have been sufficient.

\item \textbf{Prevent($e_1,e_2$)} or \textbf{Prevent($e_2,e_1$)} -- Causal relations between biomedical entities are of these classes if the emergence, application or increase of a \textbf{single or composite entity} \textit{exclusively leads} to the \textbf{eradication, prevention} or \textbf{decrease} of one or a set of entities. 

This class includes the scenario of \textit{\textbf{preventing}} a disease or condition from occurring as well as \textit{\textbf{curing}} a disease or condition if it has occurred.

\begin{enumerate}[resume]
        \item \begin{minipage}[t]{0.38\textwidth} You were treated with \textcolor{blue}{\textbf{\textless e2\textgreater tylenol and ibuprofen\textless /e2\textgreater}} for your \textcolor{red}{\textbf{\textless e1\textgreater back pain\textless /e1\textgreater}}. \end{minipage}
    \end{enumerate}
In example (7), ``tylenol'' and ``ibuprofen'' are the two different entities used in conjunction to resolve the ``back pain''. Since the causal relation is to be identified by considering them as a \textit{composite entity}, the example is labelled as Prevent($e_2, e_1$). The annotation would have been different had these entities been considered individually.

\item \textbf{Hinder($e_1,e_2$)} or \textbf{Hinder($e_2,e_1$)} -- Causal relations between biomedical entities are of these classes if the emergence, application or increase of a \textbf{single or composite entity} \textit{leads} to the \textbf{eradication, prevention} or \textbf{decrease} of one or a set of entities in a setting \textit{\textbf{where}} a number of factors are at play and the \textit{single or composite entity} under consideration is one of the contributing factors.
    
Similar to ``Prevent'', this label also includes the scenario of \textit{\textbf{hindering}} a disease or condition from occurring as well as \textit{\textbf{curing}} a disease or condition if it has occurred.

\begin{enumerate}[resume]
    
        \item \begin{minipage}[t]{0.38\textwidth} You were treated with \textcolor{blue}{\textbf{\textless e2\textgreater tylenol\textless /e2\textgreater}} and ibuprofen for your \textcolor{red}{\textbf{\textless e1\textgreater back pain\textless /e1\textgreater}}. \end{minipage}
          
    \end{enumerate}
Example (8) is the same as example (7) except for the entities in considerations. Both the entities i.e. ``tylenol'' and ``ibuprofen'' are contributing to the resolution of ``back pain''. Since the example is considering only ``tylenol'', \textit{individually as a contributing factor in part}, it is annotated as Hinder($e_2, e_1$).

This distinction between ``Prevent'' and ``Hinder'' can be useful in scenarios such as identifying conditions that may require the use of multiple drugs for treatment.
    
\item \textbf{Other} -- We defined the ``Other'' class to annotate examples with non-causal interaction between biomedical entities. Examples of the ``Other'' class can either have no relationship between biomedical entities of interest or some other semantic relationship that's not causal. Being non-causal, the ``Other'' class doesn't have a sense of direction associated with it. 

Based on our guidelines, examples with ambiguous overall context for all the annotators, entities with indirect causal association (an entity leading to a condition which in turn affects another entity) and samples from non-causal entity-pairs in the n2c2 dataset (i.e., Form-Drug, Route-Drug, etc.) are also labelled as ``Other''. 

\begin{enumerate}[resume]
    
        \item \begin{minipage}[t]{0.38\textwidth} Patient has tried and failed \textcolor{blue}{\textbf{\textless e2\textgreater Nexium\textless /e2\textgreater}}, reporting it has not helped his \textcolor{red}{\textbf{\textless e1\textgreater gastritis\textless /e1\textgreater}} for 3 months. \end{minipage}
        
        \item \begin{minipage}[t]{0.38\textwidth} Thus it was believed that the pt's \textcolor{red}{\textbf{\textless e1\textgreater altered mental status\textless /e1\textgreater}} was secondary to \textcolor{blue}{\textbf{\textless e2\textgreater narcotics\textless /e2\textgreater}} withdrawal. \end{minipage}
        
        \item \begin{minipage}[t]{0.38\textwidth} Atenolol was held given patient was still on \textcolor{blue}{\textbf{\textless e2\textgreater amiodarone\textless /e2\textgreater}} \textcolor{red}{\textbf{\textless e1\textgreater taper\textless /e1\textgreater}}. \end{minipage} 
    \end{enumerate}

In example (9), ``Nexium'' was taken to prevent / cure ``gastritis'' but the expected effect is explicitly stated to be not observed. In example (10), the ``altered mental status'' is observed due to ``narcotics withdrawal'', however, the entity span refers only to the ``narcotics''. Example (11) is from the ``Dosage-Drug'' entity-pair of the n2c2 dataset and has no causal association between the entities. 

Therefore, these examples are annotated as ``Other''. Similarly, examples with entity-pairs from ``Form-Drug'', ``Strength-Drug'', ``Frequency-Drug'', `Route-Drug'' and ``Duration-Drug'' are also labelled as ``Other''.

\end{itemize}

To summarize, we defined annotation guidelines for nine semantic causal relations (8 Causal + Other) between biomedical entities expressed in clinical notes. Our annotated dataset has examples with both explicit and implicit causality in which entities are in the same sentence or different sentences. The final count of examples for each \textit{causal} type with direction is in Table \ref{tab:annotation_types}.

\begin{table}
    \centering
    \begin{adjustbox}{max width=0.48\textwidth}
    \begin{tabular}{c c c c}
    \toprule
         &  & Annotation & Count \\
    \midrule
         \multirow{8}{*}{Causal} &  \multirow{4}{*}{$e_1$ as agent, $e_2$ as effect} & $\mathbf{Cause(e_1,e_2)}$ & 354 \\
         & & $\mathbf{Enable(e_1,e_2)}$ & 174 \\
         & & $\mathbf{Prevent(e_1,e_2)}$ & 261 \\
         & & $\mathbf{Hinder(e_1,e_2)}$ & 154 \\
    \cmidrule{2-4}
         &  \multirow{4}{*}{$e_2$ as agent, $e_1$ as effect} & $\mathbf{Cause(e_2,e_1)}$ & 370 \\
         & & $\mathbf{Enable(e_2,e_1)}$ & 176 \\
         & & $\mathbf{Prevent(e_2,e_1)}$ & 249 \\
         & & $\mathbf{Hinder(e_2,e_1)}$ & 185 \\
    \midrule
         Other & -- & $\mathbf{Other}$ & 791 \\
    \midrule
         \textbf{Total} & & & \textbf{2714} \\
    \bottomrule
    \end{tabular}
    \end{adjustbox}
    \caption{\textit{Causal} types and their final counts}
    \label{tab:annotation_types}
\end{table}

\subsection{Inter-annotator agreement}
It's difficult to comprehend narratives expressed in clinical notes due to the need of domain knowledge, short hand used by the doctors, use of abbreviations (Table \ref{tab:abbreviations}), context spread over many sentences as well as the explicit and implicit nature of communication.
\begin{table}
    \centering
    \begin{adjustbox}{max width=0.48\textwidth}
    \begin{tabular}{c c | c c}
    \toprule
         Abbreviation & Expansion & Abbreviation & Expansion \\
    \midrule
         b/o & because of &
         d/c'd & discontinued  \\ 
         HCV & Hepatitis C Virus &
         abx & anti-biotics  \\
         DM & Diabetes Mellitus  &
         c/b & complicated by \\
         s/p & status post &
         h/o & history of \\
    \bottomrule
    \end{tabular}
    \end{adjustbox}
    \caption{Clinical abbreviations in the dataset}
    \label{tab:abbreviations}
\end{table}

Three authors of this paper (all with fluency in English language and computer science background) annotated the dataset. Given the nature of our base data (MIMIC-III discharge summaries) and the critical importance of our task (causal relations between biomedical entities), the annotators followed the provided guidelines, referred to sources such as websites of Centers for Disease Control and Prevention (\textit{CDC\footnote{\url{https://www.cdc.gov/}}}), National Institute of Health (\textit{NIH\footnote{\url{ https://www.nih.gov/}}}), and  \textit{WebMD\footnote{\url{https://www.webmd.com/}}} to understand domain-specific keywords or abbreviations, and had regular discussions about the annotation tasks. 

We performed three rounds of annotation, refining our guidelines after each round by discussing various complex examples and edge cases.  We achieved an inter-annotator agreement (IAA) \textbf{Fleiss' kappa} ($\kappa$) score of $0.72 $, which indicates substantial agreement and the quality of our annotation guidelines. 

We did majority voting over the three available annotations to obtain the final gold annotations for our ``MIMICause'' dataset. In case of disagreements, another author of this paper acted as a master annotator, making the final decision on annotations after discussion with the other three annotators. 


A direct comparison of our IAA score with other works is not possible due to differences in the number of annotators, annotation labels, guidelines, reported metrics etc. for different datasets. However, for reference, we discuss IAA scores reported for the task of semantic link annotations, particularly those where $\kappa$ scores were reported. Of note is the work by Mostafazadeh et al. (\citeyear{Mostafazadeh2016CaTeRSCA}) and their annotation framework CaTeRS for temporal and causal relations in ROCStories corpus where the final $\kappa$ score achieved was 0.51 among four annotators. Similarly, Bethard et al. (\citeyear{Bethard2008LearningSL}) reported a $\kappa$ score of 0.56 and an F-measure (F-1 score) of 0.66 with two annotators labelling for only two relations viz. causal and no-rel. In the clinical domain, Bethard et al. (\citeyear{BethardSemEval2017T12}) reported a final IAA agreement (F-1) score of 0.66 on the latest Clinical TempEval dataset (Task 12 of SemEval-2017) labelled by two annotators. However, the relation types in Clinical TempEval are temporal and not causal, making the agreement score incomparable.
\section{Problem definition and Experiments}
\label{sec:experiments}
We defined our task of causality understanding as the identification of semantic causal relations between biomedical entities as expressed in clinical notes. We have a total of 2714 examples annotated with these 9 different classes (8 causal and 1 non-causal).

\subsection{Problem Formalization}
We pose the task of causal relation identification as a multi-class classification problem $f: (X, e_1, e_2) \mapsto y$, where $X$ is an input text sequence, $e_1$ and $e_2$ are the entities between which the relation is to be identified, and $y \in \mathcal{C}$ is the label from the set of \textit{nine} relations. These samples are taken from the \textit{MIMICause} dataset $\mathcal{D} = \{ \left( X, e_1, e_2, y \right)_m \}_{m=1}^{m=N}$, where $N$ is the total number of samples in the dataset. The text and entities are mathematically denoted as:
\begin{align}
    X &= [x_1, x_2, \ldots, x_{n-1}, x_n] \\
    e_1 &= X[i:j] = [x_i, x_{i+1}, \ldots, x_j] \\
    e_2 &= X[k:l] = [x_k, x_{k+1}, \ldots, x_l]
\end{align}

where $n$ is the sequence length, $i, j, k$ and $l \in \left[ 1 .. n \right]$, $i \leq j$ and $k \leq l$ i.e. entities are sub-sequences of continuous span within the text $X$. Additionally, $j < k$ or $l < i$ holds i.e. the entities $e_1$ and $e_2$ are non-overlapping and either of these can occur first in the sequence $X$.

\begin{figure*}[!h]
\centering
\begin{minipage}{.5\textwidth}
  \centering
  \includegraphics[width=\linewidth,keepaspectratio]{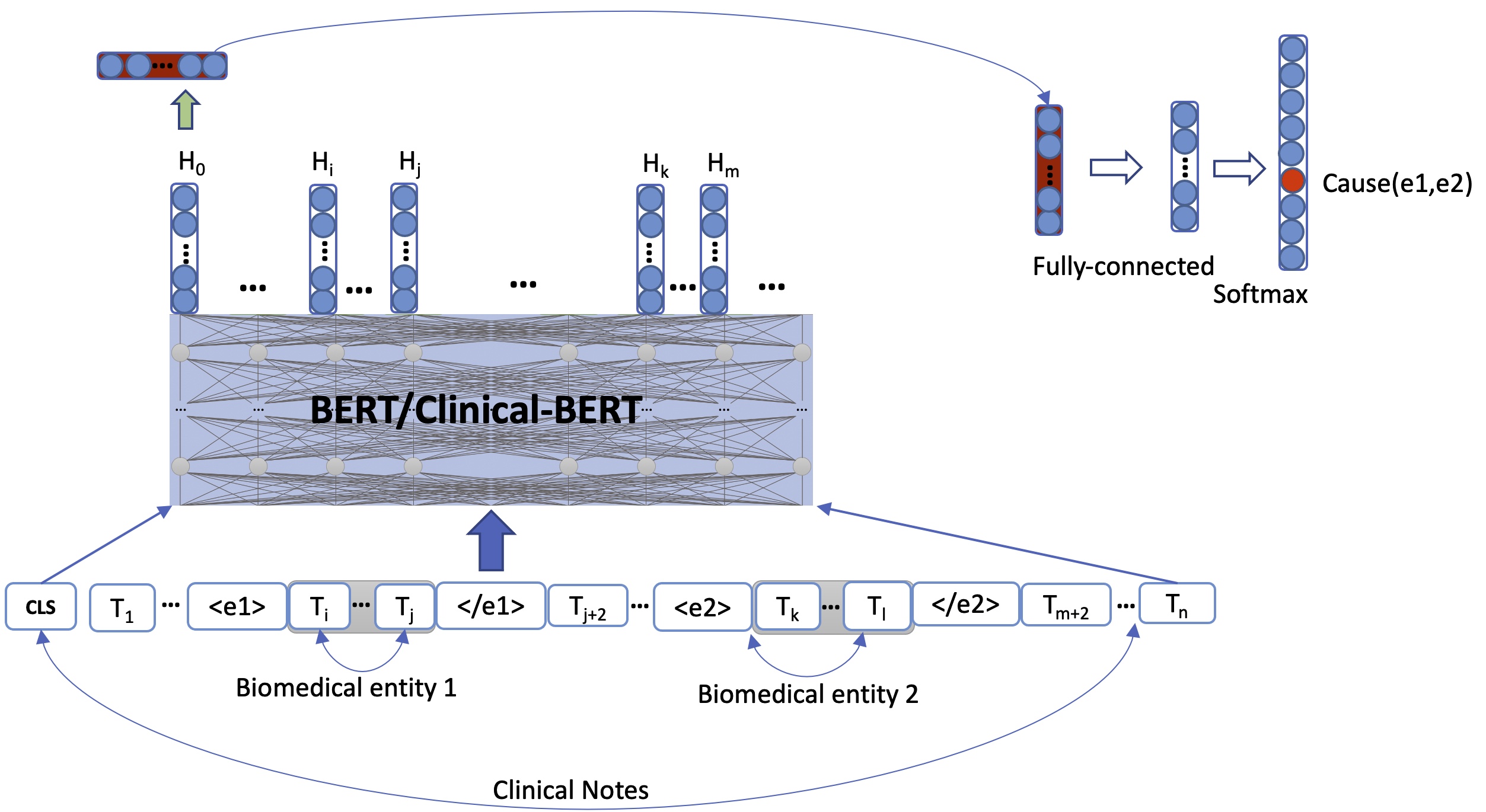}
  \caption{BERT/Clinical-BERT: FFN }
  \label{fig:ffn}
\end{minipage}%
\begin{minipage}{.5\textwidth}
  \centering
  \includegraphics[width=\linewidth,keepaspectratio]{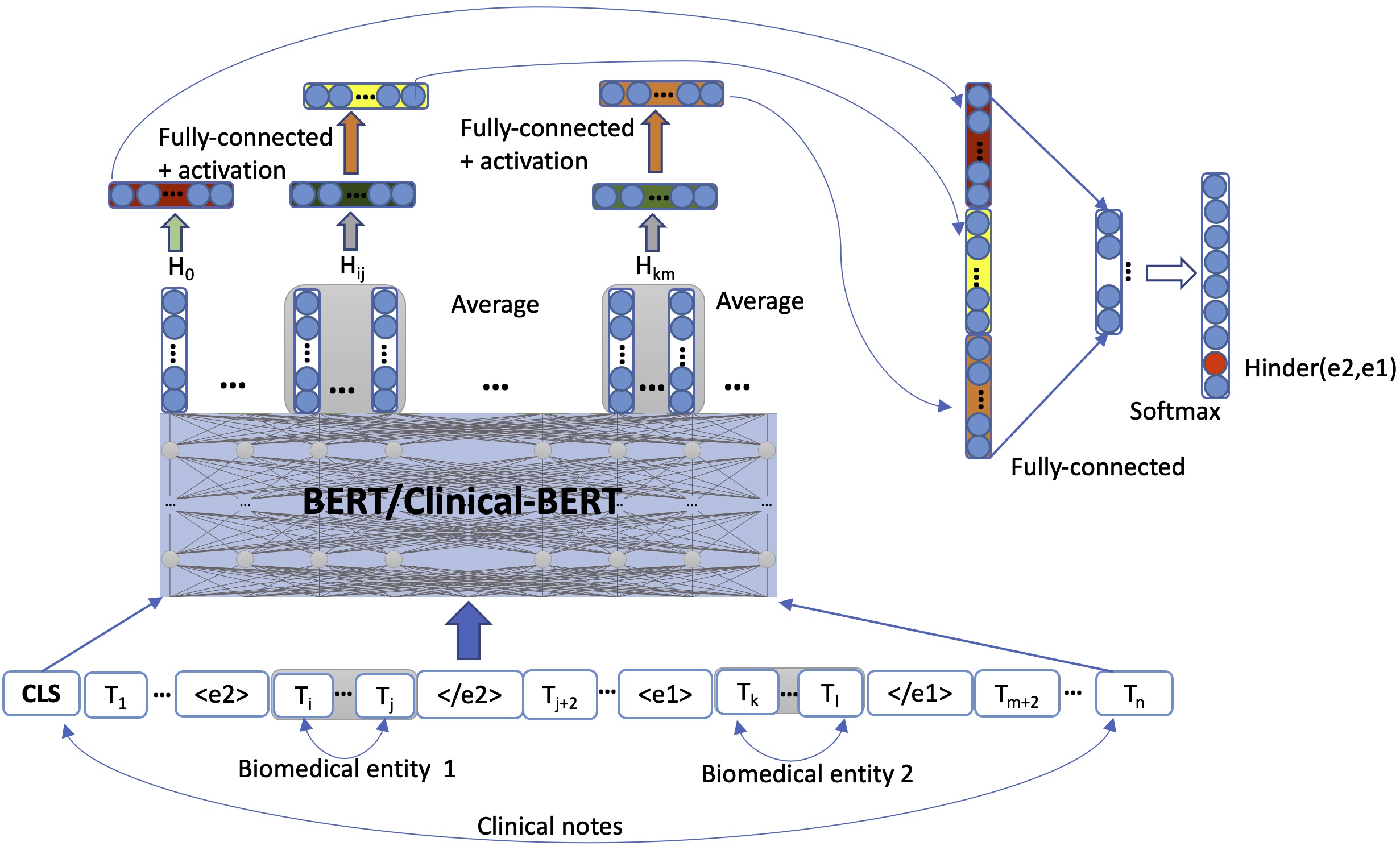}
  \caption{BERT/Clinical-BERT: FFN with entity context}
  \label{fig:entity_context}
\end{minipage}
\end{figure*}

\subsection{Models}
As a baseline for this dataset, we built our causal relation classification models using two different language models\footnote{
We use the implementation of all the encoders from the huggingface \citep{wolf-etal-2020-transformers} repository} as text encoders (BERT-BASE and Clinical-BERT) and a fully connected feed-forward network (FFN) as the classifier head. The encoder output that captures the bi-directional context of the input text \unboldmath $X$ through the [CLS] token is denoted by $H_0 \in R^d$, where $d = 768$ is the dimension of the encoded outputs from BERT-BASE / Clinical-BERT. The formulations of the layers of the classifier head are given by:
\begin{align}
    K_1 &= \textit{dropout}(ReLU(W_1 H_0 + b_1))) \\
    K_2 &= W_2 K_1 + b_2 \\
    p & = \textit{softmax}(K_2)
\end{align}


where $W_1 \in R^{d' \times d}$, $W_2 \in R^{L \times d'}$, $d'$ was set to 256 and $L = 9$ is the number of labels.

Architectures with additional context introduced between the encoder and classifier head by concatenating averaged representation of the two entities and encoder output were also tried, which led to improved results. The augmented context is denoted by:
\begin{align}
    H_{e_1} &= \frac{1}{j-i+1} \sum_{t=i}^j H_t \\
    H_{e_2} &= \frac{1}{l-k+1} \sum_{t=k}^l H_t \\
    H' &= \textit{concat}(H_0, H_{e_1}, H_{e_2}) \\
    H_0 &= \textit{dropout}(ReLU(W_0 H' + b_0))
\end{align}

\noindent where $i, j, k$ and $l$ are the start and end indices of the entities, $H_t \in R^d$, $H' \in R^{3d}$, $W_0 \in R^{d \times 3d}$ and the augmented context is assigned back to $H_0$ for feeding into the classifier head. The architecture details without and with the entity context augmentation are shown in Figure (\ref{fig:ffn}) and (\ref{fig:entity_context}) respectively. An overview of the models is given below:

\begin{itemize}
    \item \textbf{\textit{Encoder (BERT-BASE / Clinical-BERT) with feed-forward network (FFN)}} -- The overall architecture as shown in Figure \ref{fig:ffn} is a simple feed-forward network built on top of a pre-trained encoder. The input sentence is fed as a sequence of tokens to the encoder, with encoder based special tokens such as [CLS] and entity tagging tokens such as $\textless e1\textgreater, \textless /e1\textgreater$. The overall sentence context is passed through the fully connected feed-forward network to obtain class probabilities as formulated in equations (4)--(6).
    
    In addition to the BERT-BASE encoder, we also used the Clinical-BERT encoder to obtain the contextualised representation of our input examples. While BERT is pre-trained on standard corpus such as Wikipedia, Clinical-BERT is pre-trained on clinical notes and provides more relevant representation for our dataset, and hence led to a significant increase in the evaluation metrics. 
    
    \item \textbf{\textit{Encoder (BERT-BASE / Clinical-BERT) with entity context augmented feed-forward network (FFN)}} -- The overall architecture is shown in Figure \ref{fig:entity_context}. While the input with special tokens, encoding and classifier head remains the same as discussed earlier, the current architecture also enriches the sentence context with both the entities' context as formulated in equations (7)--(10). The special tokens around the entities ($\textless e1\textgreater$, $\textless /e1\textgreater$, $\textless e2\textgreater$, and $\textless /e2\textgreater$) are used to identify the tokens related to the individual entities which are then used to obtain the averaged context vector for each entity. These are then concatenated with the overall sentence context and are fed to a fully connected feed-forward network to predict the type of causal interaction expressed in the text.
    
    Similar to our previous discussion, in addition to the BERT-BASE encoder, a pre-trained Clinical-BERT encoder was also used which resulted in the highest evaluation metrics.
\end{itemize}


\begin{table}[!h]
    \centering
    \begin{adjustbox}{max width=0.48\textwidth}
    \begin{tabular}{c c c c}
    \toprule
       & Test & Val & Train \\
    \midrule
        BERT+FFN & 0.23 & 0.25 & 0.29 \\
        Clinical-BERT+FFN & 0.27 & \textbf{0.31} & 0.34 \\
        BERT+entity context+FFN & 0.54 & 0.27 & 0.56 \\
        Clinical-BERT+entity context+FFN & \textbf{0.56} & 0.30 & \textbf{0.70} \\
    \bottomrule
    \end{tabular}
    \end{adjustbox}
    \caption{Macro F1 score on test, val and train dataset}
    \label{tab:results}
\end{table}


\subsection{Results and analysis}
We trained all our models on a varied set of hyper-parameters and chose the best model from training epochs based on the maximum F1 score on the validation set. For BERT+FFN model, we achieved the best scores with a batch size of 128 and a learning rate of 5e-5. The other three models achieved reported scores with a batch size of 32 and a learning rate of 1e-3. All the models were trained until convergence with the early stopping of 7 epochs with no decrease in validation loss. We used AdamW optimizer with cross-entropy loss for all models. 

Table \ref{tab:results} shows performance measures of various models on train/val/test set. Using only the BERT-BASE encoder for the relation identification doesn't yield high scores but concatenating entity context to the BERT's encoded sentence output resulted in significant improvement. Using Clinical-BERT as base encoder resulted in additional improvements, and combining entity contexts with Clinical-BERT as base encoder resulted in the highest F1 score. While Clinical BERT was trained on the MIMIC dataset and might have seen input sequences in the test dataset, it has not seen newly defined causal classes for those sequences.   

\section{Conclusion}
In this work, we proposed annotation guidelines to capture the types and direction of causal associations, annotated a dataset of 2714 examples from de-identified clinical notes and built models to provide a baseline score for our dataset.

Even with the inherent complexities in clinical text data, following the meticulously defined annotation guidelines, we achieved a high inter-annotator agreement, i.e., Fleiss' kappa ($\kappa$) score of 0.72. Building various network architectures on top of language models, we achieved a macro F-1 score of 0.56.

An end-to-end NLP pipeline built with models for patients' data de-identification, biomedical entity extraction, and causal relations identification between various biomedical entities will be instrumental in narrative understanding from clinical notes. In the future, we are planning to extend our annotation guidelines to jointly annotate temporal and causal relations to capture the ordering of various causal interactions between biomedical entities over time. 

\section*{Acknowledgements}

We would like to thank Prof. Byron Wallace for helpful discussions and feedback.

\bibliography{MIMICause}
\bibliographystyle{acl_natbib}

\end{document}